\documentclass[final,5p,times,twocolumn]{elsarticle}

\usepackage{amsmath,amssymb,amsfonts}
\usepackage{graphicx}
\usepackage{textcomp}
\usepackage{xcolor}
\usepackage{natbib}
\usepackage{verbatim}
\usepackage{subfiles}
\usepackage{stfloats}
\usepackage{float}
\usepackage{multirow}
\usepackage{pgfplots}

\usepackage{enumitem}
\usepackage{booktabs}

\usepackage{algorithm,algorithmic}
\usepackage{caption}

\pgfplotsset{compat=1.18}

\begin{document}

\begin{frontmatter}

\title{OpenECAD: An Efficient Visual Language Model for Editable 3D-CAD Design}

\author[label1]{Zhe Yuan\corref{cor1}} %
\ead{yuanzhe1999@outlook.com}
\author[label1,label2]{Jianqi Shi} %
\author[label1,label2]{Yanhong Huang} %
\affiliation[label1]{organization={Software Engineering Institute, East China Normal University},%
            state={Shanghai},
            country={China}}
\affiliation[label2]{organization={National Trusted Embedded Software Engineering Technology Research Center},%
            state={Shanghai},
            country={China}}

\cortext[cor1]{Corresponding author.}

\begin{abstract}

Computer-aided design~(CAD) tools are utilized in the manufacturing industry for modeling everything from cups to spacecraft. These programs are complex to use and typically require years of training and experience to master. Structured and well-constrained 2D sketches and 3D constructions are crucial components of CAD modeling. A well-executed CAD model can be seamlessly integrated into the manufacturing process, thereby enhancing production efficiency. Deep generative models of 3D shapes and 3D object reconstruction models have garnered significant research interest. However, most of these models produce discrete forms of 3D objects that are not editable. Moreover, the few models based on CAD operations often have substantial input restrictions. In this work, we fine-tuned pre-trained models to create OpenECAD models (0.55B, 0.89B, 2.4B and 3.1B), leveraging the visual, logical, coding, and general capabilities of visual language models. OpenECAD models can process images of 3D designs as input and generate highly structured 2D sketches and 3D construction commands, ensuring that the designs are editable. These outputs can be directly used with existing CAD tools' APIs to generate project files. To train our network, we created a series of OpenECAD datasets. These datasets are derived from existing public CAD datasets, adjusted and augmented to meet the specific requirements of vision language model~(VLM) training. Additionally, we have introduced an approach that utilizes dependency relationships to define and generate sketches, further enriching the content and functionality of the datasets.
\end{abstract}

\begin{keyword}
small language model; visual language model; computer aided design; geometric deep learning
\end{keyword}

\end{frontmatter}

\section{Introduction\label{section:introduction}} 

In today’s digital era, computer-aided design~(CAD) tools are employed in various industrial fields for 3D shape design, including automotive, aerospace, manufacturing, and architectural design. However, due to drafting conventions and the requirements for shape constraints and edit-ability, 3D shapes are still based on 2D sketches. This approach allows for the meticulous development, relation, and annotation of all design details, mirroring the precision of traditional draftsmen. A typical 3D part drawing process involves multiple ``Sketch-Extrusion'' steps, where one sketch can correspond to multiple extrusions. A sketch consists of multiple closed loops formed by several lines and includes constraints within or between them to ensure the sketch is fully defined. The extrusion operation generates a 3D feature based on the 2D sketch(es). Generally, there is a sequential relationship between ``Sketch-Extrusion'' steps, as a sketch's reference plane may rely on an existing face in the model, and constraints within the sketch may depend on points or lines already present in the model.

There has been extensive research on 3D model generation. Most of this research focuses on creating computer-discretized forms of 3D shapes, such as 3D point clouds\cite{achlioptas2018learning, yang2018foldingnet, mo2019structurenet, yang2019pointflow, cai2020learning}, voxelized shapes\cite{girdhar2016learning, wu2016learning, li2017grass, liao2018deep}, polygon meshes\cite{groueix2018papier, wang2018pixel2mesh, nash2020polygen}, and levelset fields\cite{chen2019learning, mescheder2019occupancy, park2019deepsdf, chen2020bsp, wu2020pq}. This approach focuses on the application of precise 3D point cloud models in cyberspace, neglecting the essence of 3D shape design—the drawing process, and therefore is not suitable for the design stage of 3D models. With the development of neural networks, some studies have emerged focusing on generating 3D shapes based on points and surfaces. SolidGen\cite{jayaraman2022solidgen} and ComplexGen\cite{guo2022complexgen} generate B-rep models, while Point2Cyl\cite{uy2022point2cyl}, DeepCAD\cite{wu2021deepcad}, and Free2CAD\cite{li2022free2cad} generate CAD command sequences. In practical design workflows, generated 3D models often do not fully meet design requirements and need modifications due to production processes like draft angles and forging. Without the drawing process, these modifications can be less efficient than manually modeling from scratch, making these methods challenging to apply. Additionally, these models often have input limitations, such as requiring existing 3D point clouds or detailed hand-drawn information in isometric views.

Now, with the development of language models and multimodal language models, these limitations are being overcome. Multimodal language models can flexibly accept various forms of input and understand their meaning, thereby constraining the output. Currently, multimodal language models already have the basic ability to accept 2D views or language descriptions of 3D shapes and use CAD operations to draw them. For example, GPT-4o can possess simple 3D shape generation and understanding capabilities, CAD-LLM\cite{wu2023cad} can manipulate engineering sketches by fine-tuning pre-trained language models to generate parametric CAD.

\begin{figure}[t]
    \centering
    \includegraphics[width=0.48\textwidth]{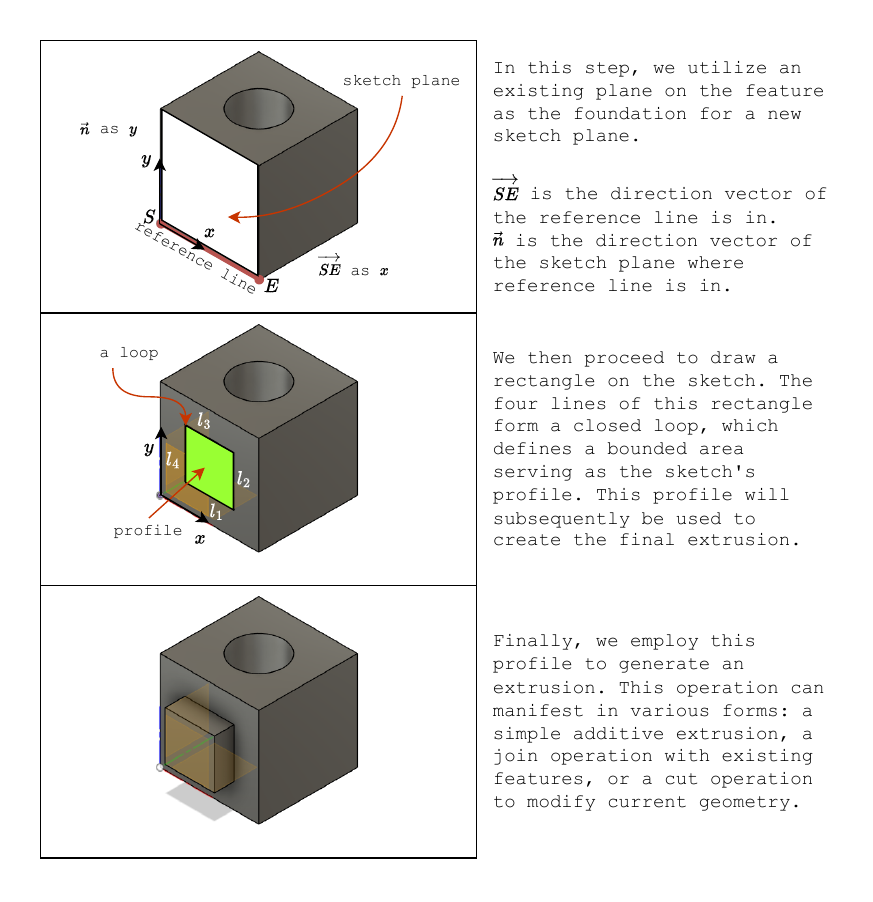}
    \caption{Example of Extrusion Feature Addition Using an Existing Face as Sketch Reference Plane.} %
    \label{fig:processofdrawing3dshape}
\end{figure}

However, current models and datasets often define all points, lines, and surfaces using coordinates within space, such as DeepCAD. This method lacks the dependencies between elements, for instance, a sketch's reference plane can be defined based on existing points, lines, and surfaces rather than solely relying on the origin and direction vectors. This dependency-based approach to defining 3D shapes, which aligns with human modeling methods, enhances robustness, readability, and editability. The Figure~\ref{fig:processofdrawing3dshape} illustrates an example of adding an extrusion feature to an existing geometry, where the new extrusion is created using an existing face as the reference plane for the sketch. Additionally, the vision towers of current multimodal language models, such as OpenAI's CLIP\cite{radford2021learning} and Google's SigLIP \cite{zhai2023sigmoid}, tend to focus on macroscopic image analysis, making it difficult to extract fine details, which leads to a significant loss of shape details during generation.

In this work, we leverage the visual, logical, coding, and general capabilities of visual language models to make the following contributions:

\begin{enumerate}
\item We established OpenECAD datasets. We transform the DeepCAD dataset into pairs of ``image-CAD operation code''. Where feasible, we employ reference planes to establish new sketch planes, moving beyond the sole reliance on points and direction vectors. The dimensions within the sketches will be consistent with the absolute space, rather than being scaled as in DeepCAD. These changes enhance the editability of the code. Additionally, we will utilize rendering tools to generate images including single random view, single isometric and transparent view, and a combination of isometric view and three orthographic views, enhancing the versatility of model inputs.

\item We trained OpenECAD models using existing small language models and multimodal small language models, including OpenELM\cite{mehtaOpenELMEfficientLanguage2024}, Gemma\cite{team2024gemma} and Phi-2\cite{abdin2024phi}. We employed TinyLLaVA's \cite{zhou2024tinyllava} training methodology, using OpenAI's CLIP or Google's SigLIP as vision tower, and training these models with the LLaVA dataset\cite{liu2024visual} to imbue them with multimodal capabilities. Subsequently, we fine-tuned both the language models and vision towers of these architectures with OpenECAD datasets and evaluated their performance in CAD design. These models, with the largest containing only 3.1B parameters, demonstrate efficient inference capabilities in single GPU environments.

\item We designed a program that can read the generated CAD code and interface with the APIs of PythonOCC, enabling the reconstruction of 3D models from the code and testing the models. Unlike previous models, OpenECAD models require only one or few 2D view(s) to understand 3D shapes, without needing point clouds or Brep files. The generated code is highly editable and aligns with human operations in CAD tools, enabling flexible manual modifications.
\end{enumerate}

\section{Related Work\label{section:related_work}}
\subsection{Sketch-based 3D Modeling Techniques}

In recent years, researchers have made significant progress in the field of automatic conversion from 2D sketches to 3D models. ``From sketches to CAM models''\cite{plumed2013sketches} proposed a method to directly generate 3D CAM models from 2D sketches by identifying high-level geometric information such as ``steps'' and ``pockets'' in 2D line drawings to construct 3D model trees. This approach simplifies traditional CAD/CAM workflows, enabling non-professionals to design and manufacture models. ``Extracting datums to reconstruct CSG models''\cite{plumed2022extracting} focused on extracting datums from 2D engineering sketches and reconstructing Constructive Solid Geometry (CSG) models through feature extraction, feature tree construction, and datum definition, effectively translating design intent. ``Isometric Conversion of Mechanical Sketches''\cite{tanaka2020isometric} introduced the SFBCM (Sketch Feature-Based Conversion Method), utilizing human perception concepts to convert complex mechanical sketches by progressively detecting and extracting simple features. Lastly, ``Method to Automatically Convert Sketches of Mechanical Objects''\cite{tanaka2020method} further expanded the types of sketches that can be processed, defining various sketch faces and features, and proposing a detailed algorithmic process that successfully handled sketches of multiple complex mechanical objects. These studies collectively advanced the development of sketch-based 3D modeling techniques, demonstrating the feasibility of extracting complex 3D features solely from 2D images. Furthermore, they laid a solid theoretical and technical foundation for the subsequent generation of 3D shapes using large neural network models.

\subsection{Small Language and Visual Language Models}

Small language models have recently become very popular due to the growing demand for privacy and local execution. Small language models are those that are smaller in size and require fewer computational resources, allowing them to run on local devices and ensure data privacy and security. Despite their smaller size, these models can still deliver efficient performance on specific tasks, making them suitable for resource-constrained environments. Currently available small language models include Microsoft's Phi series\cite{abdin2024phi}, Apple's OpenELM\cite{mehtaOpenELMEfficientLanguage2024}, Google's Gemma\cite{team2024gemma} and TinyLLaMA\cite{zhang2024tinyllama}, based on Meta's LLaMA\cite{touvron2023llama}.

Multimodal small language models further expand the applications of small language models. These models can process not only text data but also understand and generate data in various modalities, such as images and audio. By integrating information from multiple modalities, these models excel in a wider range of tasks. For example, they can convert image descriptions into text or transform hand-drawn sketches into structured textual information. Currently available multimodal small language models include the TinyLLaVA series\cite{zhou2024tinyllava}, based on small language models above, and Microsoft's Phi3 Vision\cite{abdin2024phi}.

\subsection{Generative models of 3D shapes' CAD commands}

Currently, there are several models designed for generating CAD commands for 2D Sketches or 3D shapes. SketchGraphs\cite{seff2020sketchgraphs} provides a dataset of constrained 2D sketches and proposes corresponding generative models. CurveGen and TurtleGen\cite{willis2021engineering} aim to generate usable sketches by producing closed curves, ignoring non-essential constraint solving. CADL\cite{ganin2021computer} focuses on defining sketch operations as structures similar to programming languages, leveraging language models for sketch generation.

For 3D model generation, approaches can be divided into models generating B-rep and those generating CAD commands, depending on the final output. The Boundary representation (B-rep) format is the de-facto shape representation in CAD for modeling solid and sheet objects. B-rep has poorer editability compared to other formats but can accommodate simple model modifications. SolidGen\cite{jayaraman2022solidgen} and ComplexGen\cite{guo2022complexgen} can both generate B-rep models. Models generating CAD commands can directly output CAD command sequences. Point2Cyl\cite{uy2022point2cyl} can convert cylindrical 3D point clouds into CAD command sequences. DeepCAD\cite{wu2021deepcad} provides a dataset of CAD command sequences and proposes a model capable of generating sequences randomly. Free2CAD\cite{li2022free2cad} can convert hand-drawn isometric sketches into CAD command sequences. 

As these are traditional small models, they have significant input and output limitations, making integration with existing manual modeling workflows challenging. However, they offer high-quality and structurally complete datasets, which provide a foundation for generating training data for multimodal small language models. To our knowledge, there are currently no multimodal small language models specifically trained for generating CAD commands for 3D shapes.

\section{Method overview\label{section:method}}
\begin{figure*}[t]
    \centering
    \includegraphics[width=0.98\textwidth]{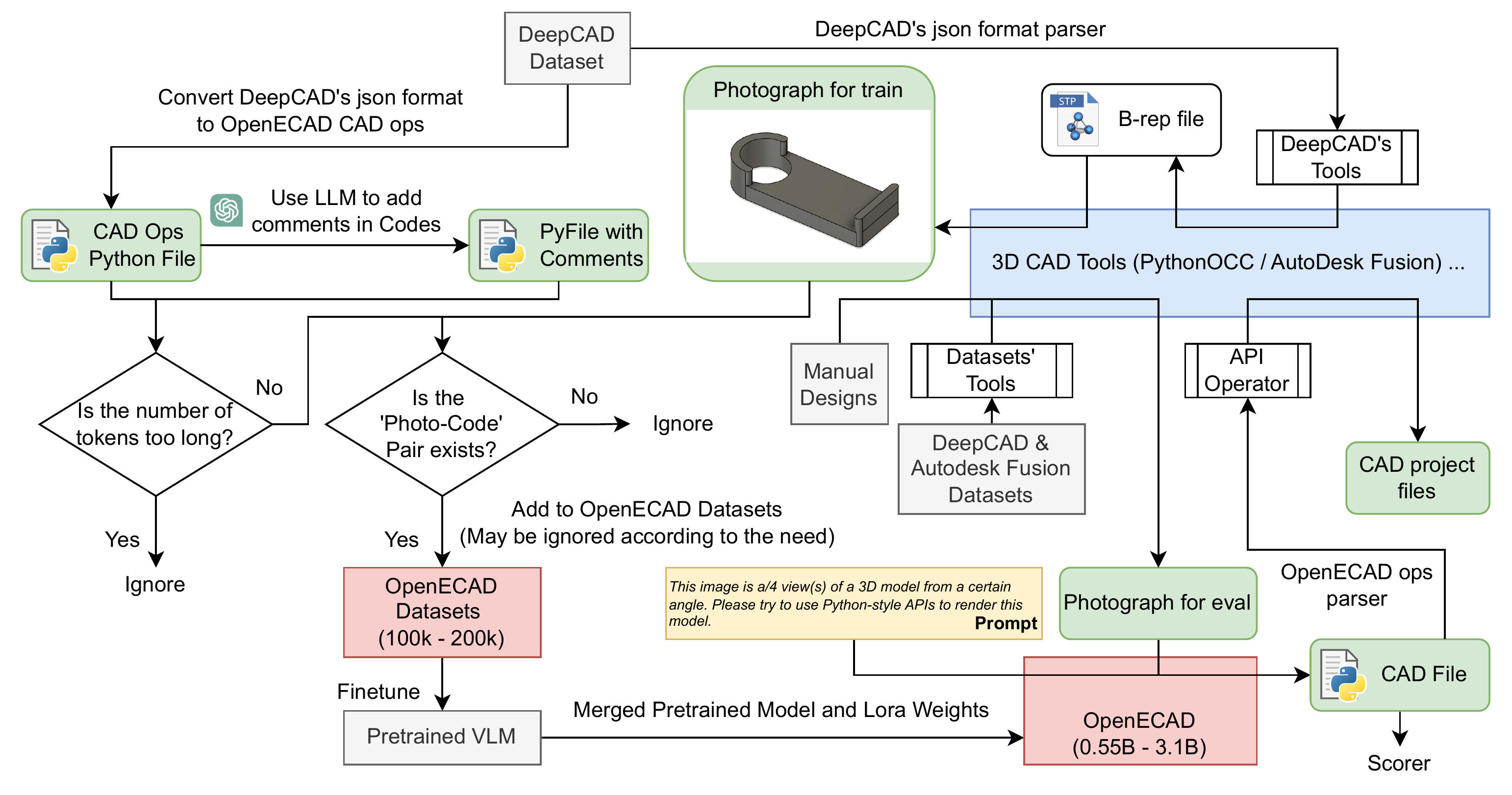}
    \caption{Overview of the OpenECAD Dataset and Model.} %
    \label{fig:method}
\end{figure*}

We now present our OpenECAD datasets and models (see Figure~\ref{fig:method}), which include a new CAD command sequence format designed for language models. These datasets are converted from the DeepCAD dataset and filtered based on the token count corresponding to the language model, resulting in multiple datasets of varying sizes. Additionally, we utilize large language models to generate high-quality natural language descriptions of OpenECAD codes. Based on multiple small language models of different sizes, we fine-tuned pre-trained vision-language models, ultimately obtaining several OpenECAD models of varying sizes. These models are compact, capable of running locally, and specifically designed for CAD applications. The python code generated by the models can produce CAD project files through corresponding API operators of CAD tools.

\section{Generation of Datasets\label{section:datasets}}
In this chapter, we encompass several key components essential for OpenECAD datasets. We cover the design of datasets (Section~\ref{section:design_datasets}), the definition of the code format for CAD operation sequences (Section~\ref{section:definition_operation}), and the translation of these sequences (Section~\ref{section:translation_cad_sequences}). Additionally, we address the generation of natural language descriptions for CAD models (Section~\ref{section:generation_descriptions}), the production of CAD model view images (Section~\ref{section:generation_images}), and the overall creation of OpenECAD datasets (Section~\ref{section:create_datasets}).

\subsection{Design of OpenECAD Datasets\label{section:design_datasets}}

CAD models can be represented in various ways. At the user interaction level with CAD software, 3D shapes are described by a series of CAD operations to create solid forms. For example, if we need to add a cuboid to an existing feature, we first need to specify the reference plane for sketching. Then, we draw a rectangle on the sketch plane. Once the sketch is completed, it forms a closed profile, and the extrude operation is applied to this profile. After the extrusion is completed, the addition of the cuboid is finished. By repeatedly performing this operation, we can create a complex 3D shape (see Figure~\ref{fig:processofdrawing3dshape}). Clearly, this CAD operation sequence can be described in natural language, as demonstrated above. Additionally, the 2D view of a CAD model is the most intuitive way to express it to humans, akin to the flowchart mentioned above.

In this work, we aim to generate CAD operation sequences using visual language models. CAD operation sequences not only interact with CAD software but also have meanings understandable by humans, allowing modifications and applications in other design processes. To constrain the generation of CAD operation sequences, we use human-understandable inputs such as model images or textual descriptions. 

\subsection{Definition of the Code Format for CAD Operation Sequences\label{section:definition_operation}}

\begin{table}[t]
\centering
\begin{tabular}{|c|c|}
\hline
Command Name & Parameters \\
\hline
\multicolumn{2}{|c|}{Curves Series} \\
\hline
add\_line & start\_point, end\_point \\
add\_arc & start\_point, end\_point, mid\_point \\
add\_circle & center\_point, radius \\
\hline
\multicolumn{2}{|c|}{Sketch and its helper} \\
\hline
add\_sketchplane & origin, normal, x\_axis \\
add\_sketchplane\_ref & extrude, origin, type, (optional values)\\
add\_profile & loops\_list \\
add\_sketch & sketchplane, profile \\
\hline
\multicolumn{2}{|c|}{Extrusion} \\
\hline
add\_extrude & sketch, operation, type, extent\_size \\ 
\hline
\end{tabular}
\caption{CAD commands and their parameters.}
\label{tab:cad_commands}
\end{table}

\begin{figure}[t]
    \centering
    \includegraphics[width=0.48\textwidth]{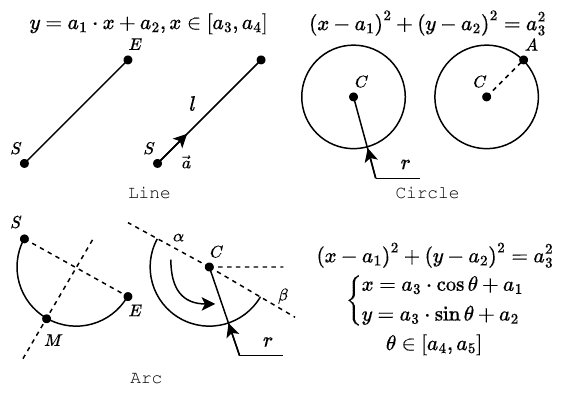}
    \caption{Comparison Diagram of Two Definitions for Line, Arc, and Circle.} %
    \label{fig:linearccircle}
\end{figure}

Mature CAD tools support a rich set of commands, most of which are simplified calls to collections of basic commands—for example, a rectangle is a collection of four line segments and some constraints. However, in practice, only a small subset of these commands is widely used. Therefore, we currently consider only the most basic and commonly used commands in our dataset (see Table~\ref{tab:cad_commands}). 

\textbf{Curve} operations include add line, arc, or circle, all of which use the most intuitive variables for definition (as shown in the definition method on the left side of Figure~\ref{fig:linearccircle}).

\textbf{Sketch} operations include defining the reference plane directly using the origin, direction vector, and x-axis direction, or using existing features to define the reference plane. When using existing features, the inputs vary depending on the type, including: using the same reference plane as the reference extrusion, using the top face or the side face obtained from the extrusion. When using the side face, the corresponding line must be specified. These methods can determine whether to reverse the direction vector or rotate the x-axis. Sketch operations also include auxiliary functions such as obtaining the profile and the sketch.

\textbf{Extrusion} operation requires inputs such as the sketch, operation type, extrusion type, and extrusion length. The operation types include simple extrusion, cut, or join with existing features, and the extrusion types include unidirectional or bidirectional extrusion.

Due to the versatility of language models, extending the dataset later is relatively easy. All operation names are chosen to be as meaningful as possible to aid the language model in understanding and using them. With these operation commands, we describe a CAD model as a sequence of executable commands in code. An example of OpenECAD code is shown in Algorithm~\ref{alg:sketch-extrusion}.

\begin{algorithm}[t]
    \caption{OpenECAD Code Example}	%
    \label{alg:sketch-extrusion}
    \begin{algorithmic}[1]
        \STATE $SketchPlane0$ = add\_sketchplane($origin$, $normal$, $x_{axis}$) 
        \STATE $Loops0$, $Curves0\_0$ = [], [] 
        \STATE $Line0\_0\_0$ = add\_line($start$, $end$) 
        \STATE ... 
        \STATE $Loop0\_0$ = add\_loop($Curves0\_0$) 
        \STATE $Profile0$ = add\_profile($Loops0\_0$) 
        \STATE $Sketch0$ = add\_sketch($SketchPlane0$, $Profile0$, $position$, $size$) 
        \STATE $Extrude0$ = add\_extrude($Sketch0$, $operation$, $type$, $extent$) 
        \STATE $SketchPlane1$ = add\_sketchplane\_ref($Extrude0$, $origin$, $type$= ``line'', $line$= $Line0\_0\_1$) 
        \STATE ...
    \end{algorithmic}
\end{algorithm}

\subsection{Translation of CAD Operation Sequences\label{section:translation_cad_sequences}}
According to the definitions provided in the previous section, we can convert any CAD model covered by the described operations (in Table~\ref{tab:cad_commands}) into a CAD operation sequence as defined by us. However, in practice, different CAD tools or datasets may have slight variations in their definitions of CAD operations. 

For curve operations, we determine and transform each curve-drawing operation according to its degrees of freedom to match the syntax of the OpenECAD dataset. For the DeepCAD dataset, we do not need to convert their definition methods; we only need to convert them into OpenECAD codes to train our vision language model~(VLM). However, for other datasets such as AutoDesk Fusion, we need to design scripts to calculate the corresponding relationships based on geometry (see Figure~\ref{fig:linearccircle}).

\begin{algorithm}[t]
    \caption{Find reference plane}	%
    \label{alg:findrefplane}
    \begin{algorithmic}[1]
        \STATE $t$ is target plane's normal vector
        \FORALL{existed extrudes: $E$}
            \STATE $n$ is $E$'s normal vector
            \IF{$n$ is parallel with $t$}
                \IF{two plane's distance is 0 or extent\_size of $E$}
                    \STATE reference plane found.
                    \RETURN
                \ELSE
                    \STATE $E$ can't contain reference plane.
                \ENDIF
            \ELSIF{$n$ is perpendicular with $t$}
                \FORALL{lines in sketch of $E$: $l$}
                    \IF{$l$ on the target plane}
                        \STATE reference plane found.
                        \RETURN
                    \ENDIF
                \ENDFOR
            \ELSE
                \STATE $E$ doesn't contain reference plane.
            \ENDIF
        \ENDFOR
    \end{algorithmic}
\end{algorithm}

For sketch operations, the primary focus is on defining the sketch reference plane, and the functions that assist in constructing the sketch do not require special conversion operations. For the DeepCAD dataset, for planes that can only be defined directly using the origin and vectors, we do not need to convert their definition methods, as the direction vector and x-axis can be easily calculated. We only need to convert them into OpenECAD codes. However, for reference planes that can be defined using existing features, we need to design an algorithm to find the appropriate plane. We will traverse all previously defined sketch planes and the planes obtained from extrusions (limited to flat surfaces, excluding curved surfaces) to find a suitable reference plane. This algorithm is shown in Algorithm~\ref{alg:findrefplane}.

For extrusion operations, no special conversion is required.

\subsection{Generation of Natural Language Descriptions for CAD Models\label{section:generation_descriptions}}

Describing CAD models in natural language is crucial for understanding 3D shapes. Adding these descriptions as annotations to the CAD operation sequence code can help people quickly grasp the meaning of each code segment and swiftly locate and modify the necessary parts. To construct natural language descriptions corresponding to CAD operation sequence codes, we employed existing large language models such as GPT-4o. By using prompts and examples, we added natural language annotations to the existing CAD operation sequence codes. The prompt used was:

\textit{“Please add comments to the drawing code below to indicate what shape is drawn by each section of the code. Don't add comment for every line or arc.”}

To prevent the language model from adding annotations to every CAD operation (like drawing lines or arcs), we constrained the model to add annotations by operation units, such as drawing a rectangle, etc.

\subsection{Generation of CAD Model View Images\label{section:generation_images}}

We have introduced how to define and generate CAD model codes and textual descriptions, providing text materials for VLM training. In this section, we will briefly explain how to generate image materials for VLM training, specifically by creating 2D views of CAD models. 

We first use the PythonOCC tool to write scripts that generate step files (B-Rep format) based on the CAD operation sequence codes. Then, we generate three types of 3D shape's images: Default View, Transparent View, and Orthographic Views. The Default View is a 2D view of the 3D shape generated by selecting a direction that forms a random angle with the isometric view direction vector. The Transparent View makes the 3D shape transparent to reveal features that are not visible from a certain direction. The Orthographic Views consist of a single image composed of the front view, top view, left view, and isometric view. We use PythonOCC to render these step files and generate the corresponding images.

\subsection{Creation of CAD Dataset\label{section:create_datasets}}

Based on the above method, we generated 3 OpenECAD datasets derived from the DeepCAD dataset. They contain 100,000, 150,000, and 200,000 ``image-code'' pairs, respectively, with code lengths limited to 1024 tokens, 2048 tokens, and 3072 tokens. We excluded designs that were generated incorrectly or had overly long CAD command sequences due to the limited number of tokens that small language models can accept, as well as designs used for the validation set. 

The distribution of ``Default View'', ``Transparent View'', and ``Orthographic Views'' in the datasets is shown in the Table~\ref{tab:disdatasets}. ``$x$\_directout\_$y$'' indicates that the dataset contains $x$ ``image-code'' pairs, with the code length limited to $y$ tokens. This dataset is used for directly outputting complete OpenECAD code from image input.

\begin{table*}[t]
\centering
\begin{tabular}{|c|c|c|c|}
\hline
Dataset Name & Default View & Transparent View & Orthographic Views \\
\hline
100k\_directout\_1k & 62500 & 25000 & 12500 \\
150k\_directout\_2k & 90000 & 37500 & 22500 \\
200k\_directout\_3k & 120000 & 50000 & 30000 \\
\hline
\end{tabular}
\caption{The distribution of OpenECAD datasets.}
\label{tab:disdatasets}
\end{table*}

\begin{figure}[t]
    \centering
    \includegraphics[width=0.48\textwidth]{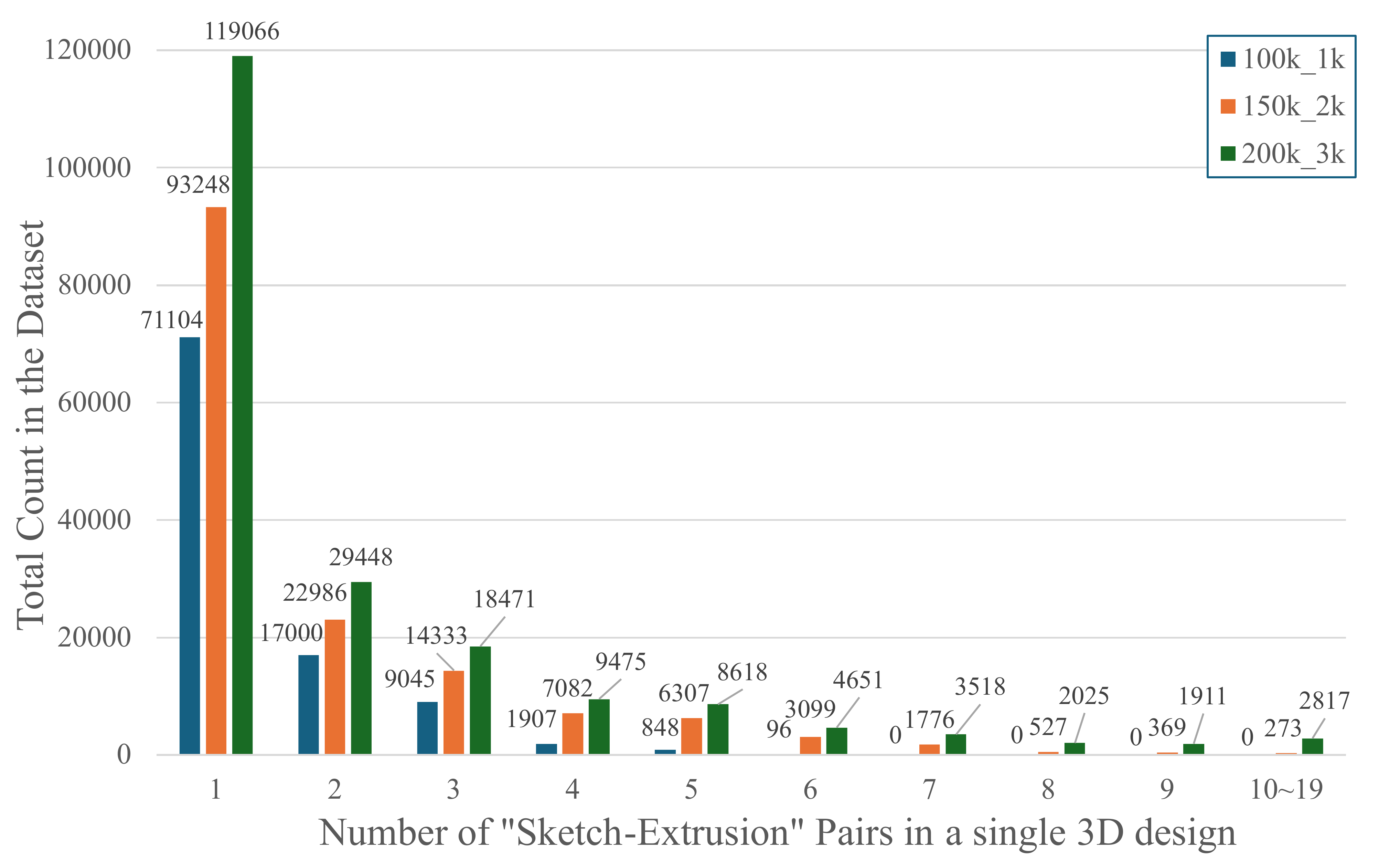}
    \caption{The statistical distribution of the number of ``Sketch-Extrusion'' steps in OpenECAD datasets.}
    \label{fig:3dstatistical}
\end{figure}

The statistical distribution of the number of ``Sketch-Extrusion'' steps in the OpenECAD datasets is shown in the Figure \ref{fig:3dstatistical}. As shown in the figure, designs with only one ``sketch-extrude'' pair still constitute the majority of datasets, especially in datasets with strict token limits. This is partly because the token limit prevents OpenECAD code from being long enough to accommodate more ``sketch-extrude'' pairs in a single design, and partly because such designs make up the majority of the DeepCAD dataset. In datasets where the token limit is less restrictive, more complex designs have a more significant proportion.

\section{Training and Fine-tuning Vision Language Models\label{section:models}}

In this chapter, we introduce the process of obtaining OpenECAD models. We discuss the choice of language models and multimodal approach (Section~\ref{section:choiceslm}), as well as the fine-tuning of a pre-trained visual language model (Section~\ref{section:finetunevlm}).

\subsection{The choice of language models and multimodal approach\label{section:choiceslm}}

Due to the high confidentiality requirements of manufacturing enterprises and the substantial CPU and GPU resources typically occupied by CAD design tools, we chose OpenELM\cite{mehtaOpenELMEfficientLanguage2024} (450M), Gemma\cite{team2024gemma} (2B) and Phi-2 (2.7B)\cite{abdin2024phi} with relatively small parameter sizes as the foundation for our language model. These models are trained on high-quality data and optimized algorithms. It is worth noting that the context acceptance length for OpenELM (450M) is 2048 tokens. In contrast, Gemma and Phi has a minimum context acceptance length of 3072 tokens, with an option to use a version supporting up to 128k tokens at the cost of some content accuracy.

To endow the language model foundation with multimodal capabilities, OpenAI's CLIP\cite{radford2021learning} and Google's SigLIP are selected \cite{zhai2023sigmoid} as visual encoders for OpenELM (450M), Gemma (2B) and Phi-2 (2.7B). These models are trained on large datasets consisting of images paired with corresponding text descriptions, combining visual and textual information to make predictions.

To train the entire model with multimodal conversational abilities, the LLaVA\cite{liu2024visual} method and its dataset are employed. LLaVA combines a visual encoder and a large language model in an innovative multimodal approach, enabling comprehensive visual and language understanding. LLaVA uses GPT-4 to generate multimodal language-image instruction data. We utilized the training framework and the pretrained models provided by TinyLLaVA\cite{zhou2024tinyllava}.

\subsection{Fine-tuning on a pre-trained visual language model\label{section:finetunevlm}}

\begin{table}[t]
\centering
\begin{tabular}{|l|c|}
\hline
\multicolumn{2}{|c|}{OpenECAD 0.55B \& 0.89B \& 3.1B} \\
\hline
CPU & Intel Xeon Platinum 8352V\\
\hline
GPU & 2 x Nvidia GeForce RTX 4090\\
\hline
RAM & 90 GB\\
OS & Ubuntu Server 22.04 LTS\\
\hline
\hline
\multicolumn{2}{|c|}{OpenECAD 2.4B} \\
\hline
CPU & Intel Xeon Platinum 8352V\\
\hline
GPU & 2 x Nvidia GeForce RTX 4080 Super (as one vGPU)\\
\hline
RAM & 90 GB\\
OS & Ubuntu Server 22.04 LTS\\
\hline
\end{tabular}
\caption{Development and running environment.}
\label{tab:environment}
\end{table}

To enable the general visual-language model to generate CAD code, we fine-tuned it using the LoRA\cite{hu2021lora} method on the pre-trained OpenELM-CLIP, OpenELM-SigLIP, Gemma-SigLIP and Phi-2-SigLIP models. LoRA stands for Low-Rank Adaptation of Large Language Models. In the field of natural language processing (NLP), there is a common paradigm that involves large-scale pre-training on general domain data followed by adaptation to specific tasks or domains. However, as we pre-train larger models, fully fine-tuning all model parameters becomes less feasible due to computational costs. LoRA proposes a novel approach. Instead of full fine-tuning, LoRA freezes the pre-trained model weights and injects trainable rank decomposition matrices into each layer of the Transformer architecture. This significantly reduces the number of trainable parameters for downstream tasks.

The specific development, training, and runtime environments are shown in the Table~\ref{tab:environment}. The loss function used was CrossEntropyLoss, and the training method was instruction fine-tuning. The learning rate was set to $10^{-4}$, and the learning rate scheduler type was cosine. The training of all OpenECAD models were conducted with a train batch size of 2 per device and gradient accumulation steps of 2. The Rank and Alpha for LoRA were set to 128 and 256. Besides using LLM to add comments to a small portion of OpenECAD datasets' codes for data augmentation, further data augmentation is of limited significance since the datasets are already sufficiently large for fine-tuning.

We performed LoRA fine-tuning on the pre-trained OpenELM-CLIP~(0.55B), OpenELM-SigLIP~(0.89B) and Phi-2-SigLIP~(3.1B) models. The training was conducted on 2 Nvidia GeForce RTX 4090 GPUs. Under the environment described in Table~\ref{tab:environment}, fine-tuning on the pre-trained OpenELM-CLIP~(0.55B) and OpenELM-SigLIP~(0.89B) models with the OpenECAD 100k\_directout\_1k dataset took approximately 2 hours, fine-tuning on the pre-trained Phi-2-SigLIP~(3.1B) model with the OpenECAD 200k\_directout\_3k dataset took approximately 20 hours.

Similarly, we used 2 Nvidia GeForce RTX 4080 Super GPUs as a single vGPU to perform LoRA fine-tuning on the Gemma-SigLIP~(2.4B) model. Under the environment described in Table~\ref{tab:environment}, fine-tuning on the pre-trained Gemma-SigLIP~(2.4B) model with the OpenECAD 150k\_directout\_2k dataset took approximately 30 hours

\begin{figure}[t]
    \centering
    \includegraphics[width=0.48\textwidth]{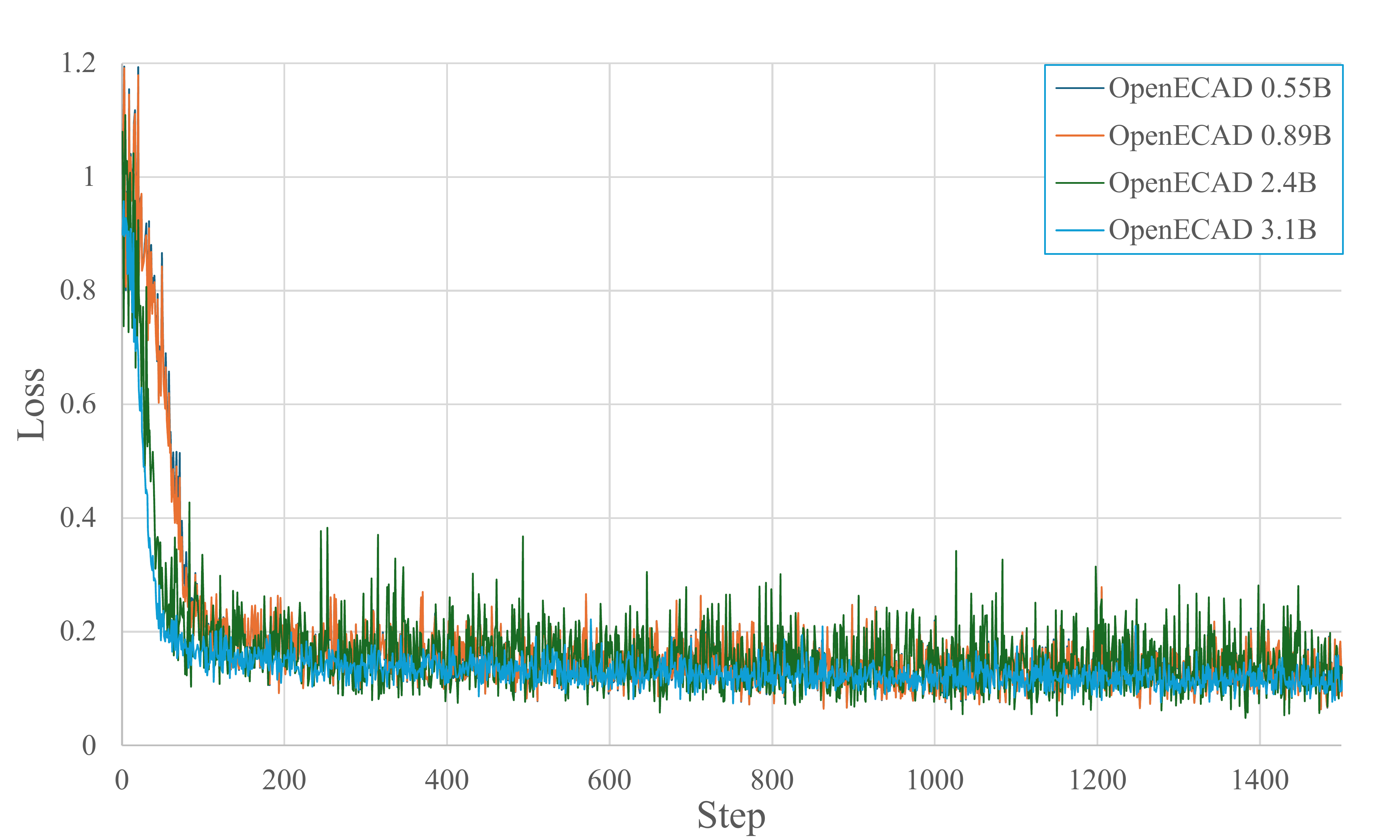}
    \caption{The partial loss curves for the OpenECAD 0.55B, 0.89B, 2.4B, and 3.1B models during training.}
    \label{fig:loss}
\end{figure}

The loss curves of the first 1500 steps for the OpenECAD 0.55B, 0.89B, 2.4B, and 3.1B models during training are illustrated in Figure~\ref{fig:loss}. As shown, the models converge after the first 1500 steps of fine-tuning.

\section{Experiments\label{section:experiments}}

In this chapter, we evaluate the ability of OpenECAD models to generate CAD designs based on input images and analyze the results. As there are no existing CAD models that use images for CAD generation, we first define evaluation metrics (Section~\ref{section:metrics}) and create CAD designs for testing (Section~\ref{section:designsfortest}). Following this, we test the models, obtain the results (Section~\ref{section:evalresult}), and analyze them (Section~\ref{section:analysis}). We also provide an example of drawing a simple table (Section~\ref{section:exampleofmodel}) and discuss future work (Section~\ref{section:future_work}).

\subsection{Evaluation Metrics\label{section:metrics}}

\begin{figure*}[t]
    \centering
    \includegraphics[width=0.98\textwidth]{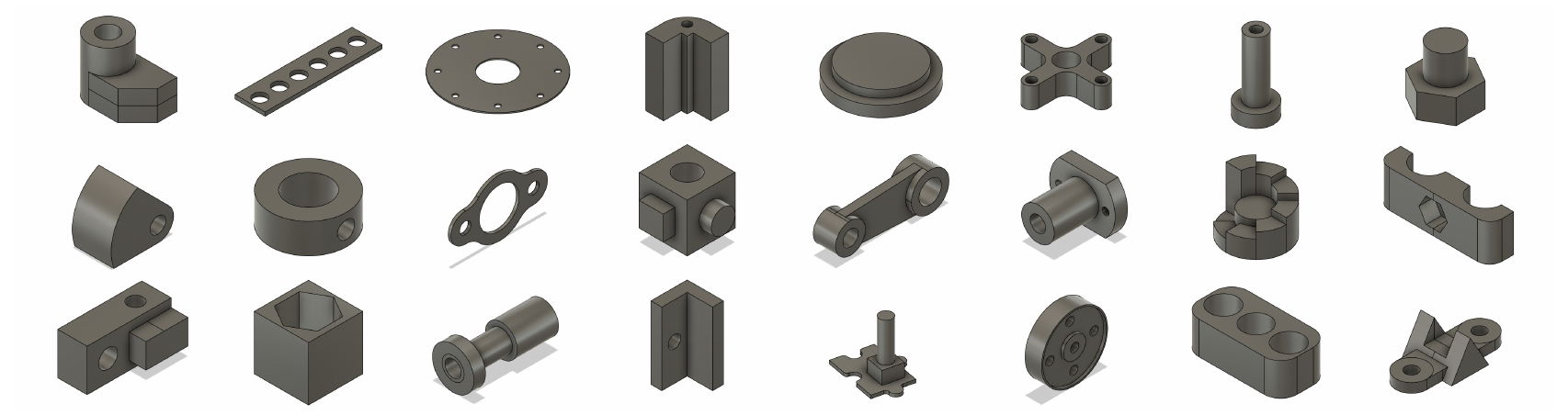}
    \caption{Example illustrations of CAD designs for evaluating.} %
    \label{fig:testexample}
\end{figure*}

For a given 3D shape, there are often multiple methods to create it using CAD tools. Moreover, the input views of the model generally lack dimensional information, meaning the generated model only adheres to the proportional relationships visible in the image. Therefore, directly comparing the generated CAD operation sequence codes is not reasonable. To address this, we designed a scoring algorithm and created several test CAD designs (see Figure~\ref{fig:testexample}). We used specific views of these designs as input images for evaluation, serving as a metric for assessing the model's generation capability.

\begin{table}[t]
\centering
\begin{tabular}{|l|c|}
\hline
Scoring Item & Score (out of 100) \\
\hline
Is the code executable? & 10 * exec\_flag \\
Accuracy of All Curves & 45 * acc \\
Accuracy of Loops count & 5 * acc \\
Loops integrated score & 40 * score \\
\hline
\multicolumn{2}{|l|}{Fo each Loop's score (out of 100)} \\
\hline
If loop is absolutely correct & 100 \\
Else calculate acc of curves & 90 * acc \\
\hline
\end{tabular}
\caption{Scoring Table.}
\label{tab:score_tab}
\end{table}

\subsubsection{Scoring Algorithm for Evaluating Outputs}

To assess the generative capability of our model, we designed a scoring algorithm. The algorithm verifies if the generated code can be executed directly by the API operator to create a model, the accuracy of curve operations, the accuracy of the number of loops, and provides a comprehensive score for the loops. We separately evaluate the loops in the 2D sketches during the modeling process to ensure precise assessment, as these loops may or may not be considered a single sketch during modeling. We score the results based on executability and the correctness of the 3D shapes (see Table~\ref{tab:score_tab}). 

For the comprehensive scoring of loops, we first check if the two loops are completely identical, meaning the types and order of the curves are exactly the same, and assign a score of 100. If not, the score is recorded as 90\% of the curve accuracy.
The specific calculation method is shown in the following formula.

\begin{equation}
score = 10e + 45acc_{c} + 5acc_{l} + \frac{40}{100L} \sum_{i=1}^{L} ( 10s_{i} + 90acc_{i} )
\end{equation}

where \( e \) denotes executability, \( acc_c \) represents accuracy of curve operations, \( acc_l \) represents accuracy of the number of loops, \( L \) represents the number of Loops, \( s_i \) and \( acc_i \) indicate whether the loop is absolutely correct and the accuracy of curve operations for the \( i \)-th loop. $e, acc_c, acc_l, s_i, acc_i \in [0, 1] $.

\subsection{CAD Designs for Test\label{section:designsfortest}}

\begin{figure}[t]
    \centering
    \includegraphics[width=0.48\textwidth]{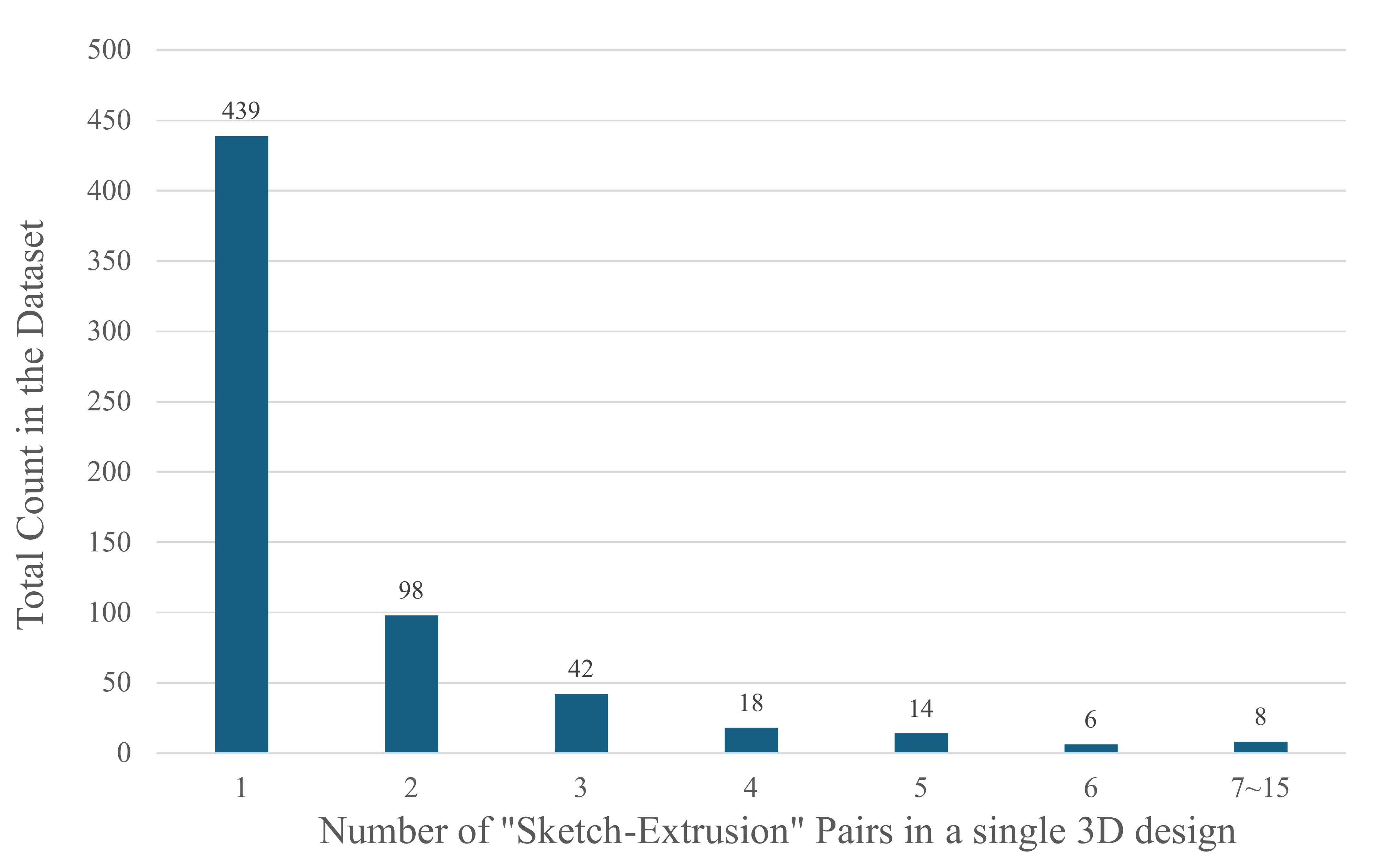}
    \caption{The statistical distribution of the number of ``Sketch-Extrusion'' steps of CAD designs for evaluating.} %
    \label{fig:3dstatisticaleval}
\end{figure}

To test the model, we require high-quality CAD designs. We extracted 625 3D designs from the OpenECAD dataset that were not included in the training set. Similar to the method used for generating the training set, we created their standard views, perspective views, and four-angle views. Some examples are shown in the Figure~\ref{fig:testexample}. The resolution of the images is 640x480. The statistical distribution of the number of ``Sketch-Extrusion'' steps in the  is shown in the Figure \ref{fig:3dstatisticaleval}.

\subsection{Evaluation Results\label{section:evalresult}}

\subsubsection{The Inference Speed of the Models}

The input was standardized to a 640x480 image of a 3D shape, with the prompt: 

\textit{``This image is a(4) view(s) of a 3D model from a certain angle. Please try to use Python-style APIs to render this model.''}

For all OpenECAD models, we used an Nvidia GeForce GTX 1080 Ti (11 GB VRAM) for inference. OpenECAD 0.55B and 0.89B models cost a peak memory usage of less than 4GB. Getting a 3D shape's code took approximately 30 seconds. For OpenECAD 2.4B, it infers with a peak memory usage of less than 9GB. OpenECAD 3.1B infers with a peak memory usage of less than 11GB. Getting a 3D shape's code took approximately 45 seconds using OpenECAD 2.4B and 3.1B models.

\subsubsection{Generating Code and Rendering}

\begin{figure*}[t]
    \centering
    \includegraphics[width=0.98\textwidth]{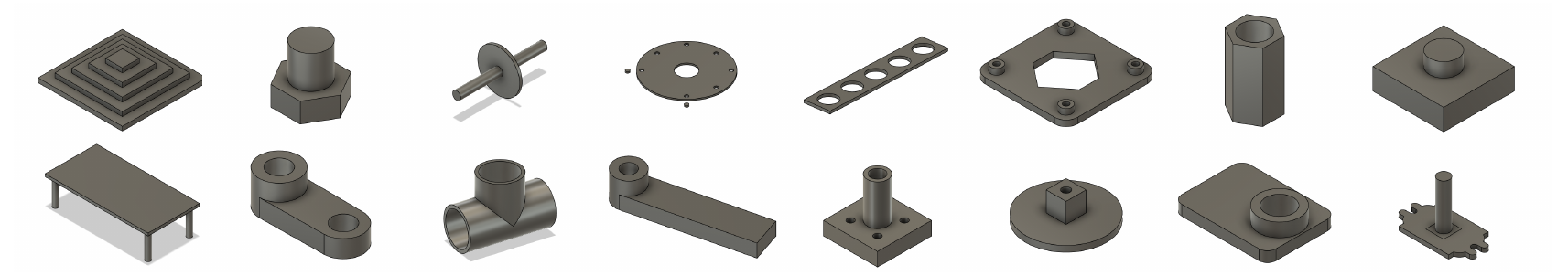}
    \caption{Some successful rendered test results illustrations} %
    \label{fig:evalresult}
\end{figure*}

We tested OpenECAD versions 0.55B, 0.89B, 2.4B and 3.1B, and also tested GPT-4o-mini (2024-07-18 version). Some successfully rendered test results are illustrated in Figure~\ref{fig:evalresult}. Since GPT-4o-mini was not trained specifically for the use case described in this paper, we used the following prompt for testing:

\textit{``I am going to provide you with a 3D shape's corresponding Python code for its modeling process. After that, I will input images of other 3D shapes, and please provide the corresponding Python code for them. $<$OpenECAD style Code Example$>$ This image is a view of a 3D model from a certain angle. Please write the codes of it in the format above.''}

During the testing of OpenECAD, the maximum number of new tokens that the 0.55B, 0.89B, 2.4B, and 3.1B models can generate is 1536, 1152, 2048, and 2048 respectively. Due to these context limitations, there are instances where the generated code is incomplete. GPT-4o-mini, due to its support for a longer context, did not encounter issues with incomplete code generation.

Similar to generating datasets, we use the PythonOCC tool to write a library that attempts to directly run the generated code to create CAD projects and output them as STEP files (B-Rep format). Then, we use PythonOCC to render the STEP files to obtain views. Since the generated programs from the model cannot ensure complete accuracy, errors may occur that the API operator cannot handle, leading to model rendering failures.

Based on the scoring method mentioned above, we obtained Curves accuracy, Loops count accuracy, Loops score, Execution score, and Overall score for each model and input type, as shown in the Table~\ref{tab:result_tab}. Additionally, we calculated the scores excluding designs with only one ``Sketch-Extrusion'' pair.

\begin{table*}[t]
\centering
\begin{tabular}{|c|c|c|c|c|c|c|c|c|c|c|}
\hline
 & \multicolumn{5}{|c|}{Default View as input} & \multicolumn{5}{|c|}{Default View without only 1 pair designs} \\ \hline
Model & 0.55B & 0.89B & 2.4B & 3.1B & gpt-4o-mini & 0.55B & 0.89B & 2.4B & 3.1B & gpt-4o-mini \\ \hline
Curves accuracy & 85.95 & 88.97 & 91.23 & 87.71 & 40.11 & 68.65 & 74.27 & 76.01 & 70.29 & 25.81 \\ \hline
Loops accuracy & 84.27 & 86.79 & 90.62 & 86.98 & 75.99 & 74.23 & 76.72 & 76.43 & 70.63 & 57.10 \\ \hline
Loops score & 77.73 & 80.03 & 84.20 & 81.46 & 34.45 & 65.50 & 68.57 & 72.96 & 69.18 & 25.07 \\ \hline
Execution score & 99.04 & 96.64 & 95.36 & 94.40 & 40.96 & 98.92 & 90.32 & 85.48 & 84.95 & 24.73 \\ \hline
Overall score & 83.89 & 86.05 & 88.80 & 85.84 & 39.72 & 70.70 & 73.72 & 75.76 & 71.33 & 26.97 \\ \hline
 & \multicolumn{5}{|c|}{Transparent View as input} & \multicolumn{5}{|c|}{Transparent View without only 1 pair designs} \\ \hline
Model & 0.55B & 0.89B & 2.4B & 3.1B & gpt-4o-mini & 0.55B & 0.89B & 2.4B & 3.1B & gpt-4o-mini \\ \hline
Curves accuracy & 84.57 & 88.03 & 90.68 & 86.72 & 47.38 & 70.00 & 73.65 & 78.52 & 70.78 & 30.00 \\ \hline
Loops accuracy & 83.12 & 84.98 & 88.74 & 85.90 & 76.20 & 75.80 & 76.49 & 76.57 & 73.36 & 58.54 \\ \hline
Loops score & 77.57 & 79.83 & 83.94 & 81.32 & 41.02 & 66.68 & 68.71 & 75.41 & 71.71 & 28.88 \\ \hline
Execution score & 99.20 & 97.76 & 95.04 & 93.44 & 44.48 & 99.46 & 93.55 & 85.48 & 82.26 & 25.81 \\ \hline
Overall score & 83.16 & 85.57 & 88.32 & 85.19 & 45.99 & 71.91 & 73.81 & 77.88 & 72.43 & 30.56 \\ \hline
 & \multicolumn{5}{|c|}{Orthographic View as input} & \multicolumn{5}{|c|}{Orthographic View without only 1 pair designs} \\ \hline
Model & 0.55B & 0.89B & 2.4B & 3.1B & gpt-4o-mini & 0.55B & 0.89B & 2.4B & 3.1B & gpt-4o-mini \\ \hline
Curves accuracy & 85.28 & 87.02 & 90.34 & 87.71 & 31.78 & 67.63 & 72.08 & 74.91 & 69.32 & 22.96 \\ \hline
Loops accuracy & 82.73 & 84.46 & 89.17 & 87.02 & 71.11 & 70.31 & 75.17 & 73.89 & 70.08 & 54.86 \\ \hline
Loops score & 76.50 & 79.02 & 83.21 & 81.14 & 26.75 & 63.18 & 68.47 & 72.15 & 68.50 & 21.97 \\ \hline
Execution score & 99.68 & 97.44 & 95.52 & 95.20 & 34.88 & 98.92 & 91.94 & 86.56 & 84.41 & 23.12 \\ \hline
Overall score & 83.08 & 84.73 & 87.95 & 85.80 & 32.04 & 69.11 & 72.78 & 74.92 & 70.54 & 24.17 \\ \hline
\end{tabular}
\caption{Evaluation Results Table.}
\label{tab:result_tab}
\end{table*}

As shown in Table~\ref{tab:result_tab}, OpenECAD 2.4B achieved the highest overall score. The 0.55B and 0.89B models, despite generating the fewest tokens, scored relatively high in executability. This might be because their inability to understand complex models leads them to finish code generation prematurely. The 3.1B model did not perform as expected, possibly due to the Phi series models having inferior geometric understanding compared to the Gemma series models. All OpenECAD models demonstrated significant advantages over the unrefined GPT-4o-mini, whose sole advantage was its ability to handle the entire code due to its extended context length.

From the perspective of using different types of view as input, using single-view transparent images as input shows a significant advantage in designs with more than one ``sketch extrusion'' pair because they contain more information. However, using four views as input (isometric view + three orthographic views) performs worse, possibly because the vision tower cannot handle so much information, or the language model cannot establish connections between them. The additional views increase the amount of irrelevant information, adding pressure to the model.

\subsection{Analysis of OpenECAD's outputs\label{section:analysis}}

\subsubsection{Code Generation Analysis}
For the generated code, the language model effectively adheres to the OpenECAD dataset's syntax, even without being explicitly instructed on the specific details of the syntax. Regarding the curves in the sketch, the language model successfully ensures that consecutive curves are connected end-to-end, with the first and last curves also joined, thus maintaining closure. Additionally, the language model adeptly manages the positions of points to ensure the profile's validity. For the fully generated code, the vast majority can be executed and rendered directly. For generated code with runtime errors, this includes both execution crashes and issues where the B-Rep file fails to render correctly.

\subsubsection{Generation Error Analysis}

In the generated results, several typical errors were observed in the Table~\ref{tab:error_analysis}.

\begin{table*}[!t]
\centering
\begin{tabular}{|l|p{9cm}|}
\hline
\textbf{Error Case (Left: target, Right: generation result)} & \textbf{Analysis} \\ \hline
\raisebox{-\totalheight}{\includegraphics[width=0.4\textwidth]{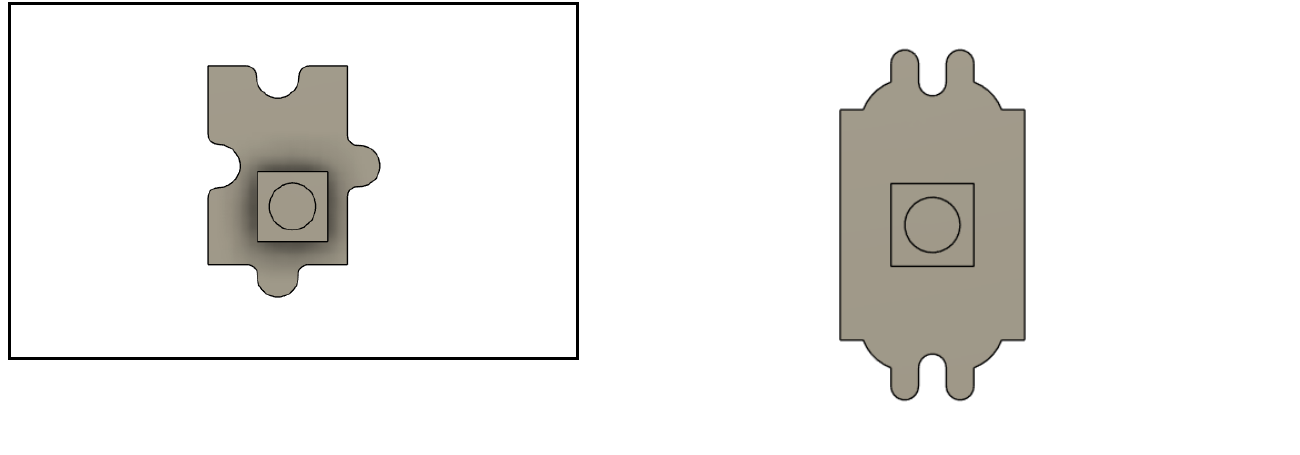}} & For complex and irregular shapes, the corresponding sketches often fail to generate correctly, or the profiles of the 2D sketches are inaccurately created. Possible reasons include the vision tower's low resolution, which hinders accurate recognition of complex shapes, and the language model's inability to comprehend complex shapes from the CAD operation code. These issues ultimately result in incorrect loops or intersecting loops, leading to erroneous sketch profiles and broken extrusions. \\ \hline
\raisebox{-\totalheight}{\includegraphics[width=0.4\textwidth]{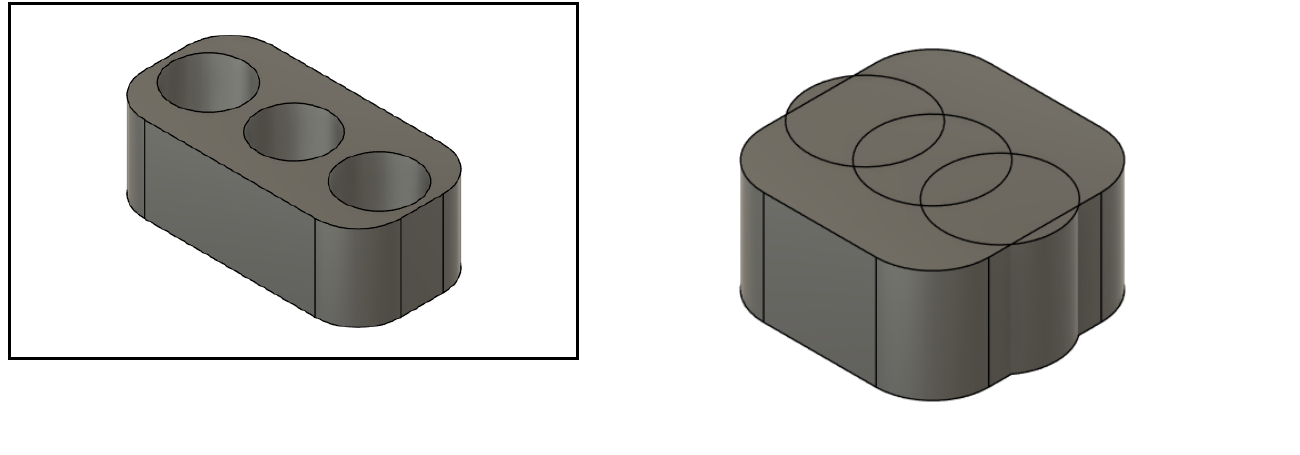}} & The model lacks spatial reasoning capabilities, resulting in incorrect handling of the size proportions between sketches and extrusions. This leads to extrusions that are either too long or too small, or sketches that are either too large or too small, causing discrepancies in size relative to other parts. The lack of numerical understanding is a common issue with language models, which can be relatively easily resolved by post-editing the extent size of the extrusions or the size of curves in the sketches in the CAD code or project. \\ \hline
\raisebox{-\totalheight}{\includegraphics[width=0.4\textwidth]{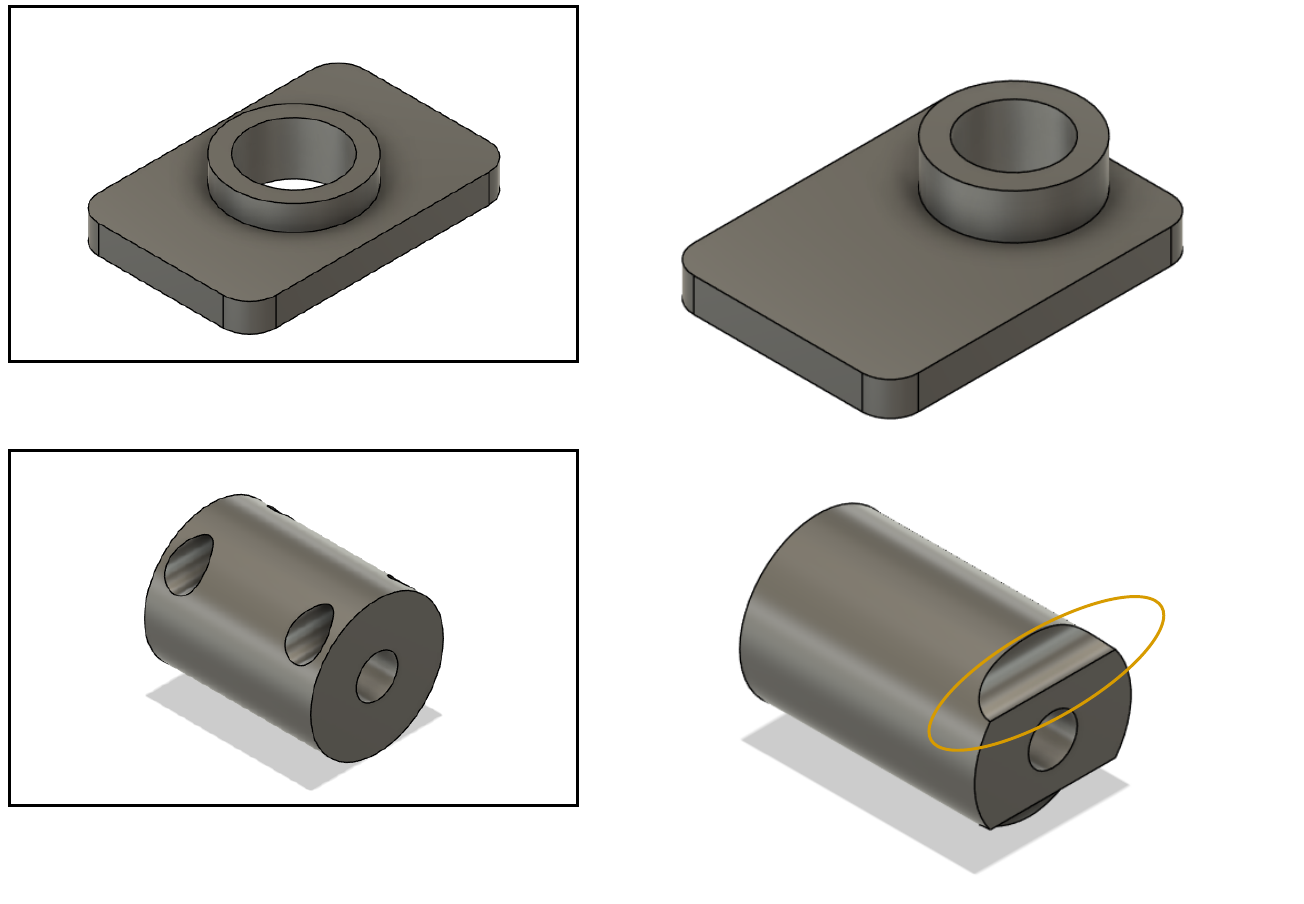}} & When multiple ``Sketch-Extrusion'' steps are involved, the positions of the sketches in subsequent steps are often incorrect, leading to errors in the final overall model. This can be resolved by post-editing the reference or absolute positions of the sketches in the CAD code or project. However, for complex models, this solution might be more challenging.
 \\ \hline
\raisebox{-\totalheight}{\includegraphics[width=0.4\textwidth]{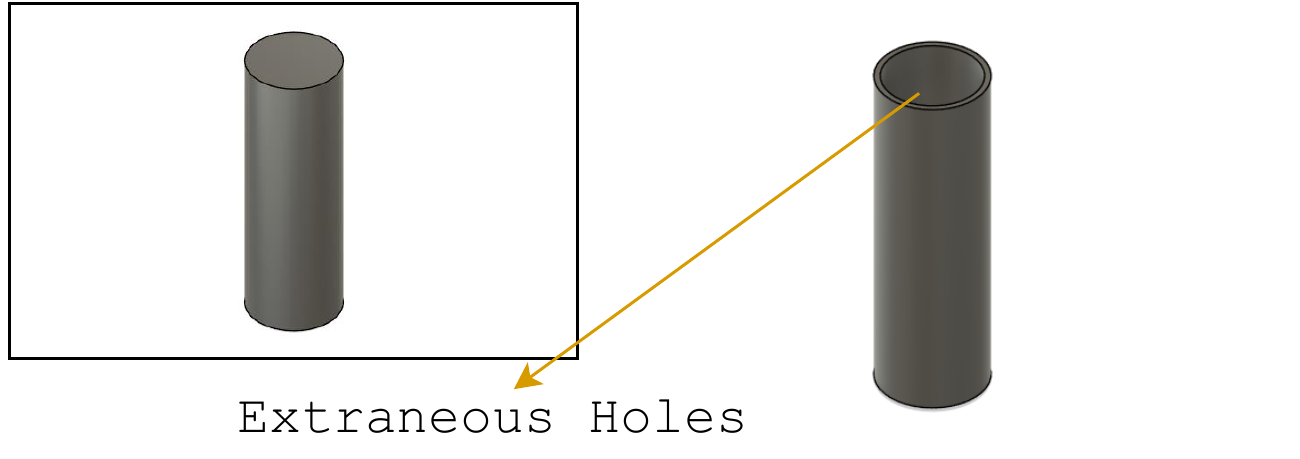}} & Due to the minimal color difference between the outer surface of a solid cylinder and the inner wall of a hollow cylinder, the primary distinguishing feature is the inner wall circle at the top. This makes it difficult for the model to differentiate between solid and hollow cylinders, leading to confusion. This issue also extends to other similar shapes with holes. \\ \hline
\raisebox{-\totalheight}{\includegraphics[width=0.4\textwidth]{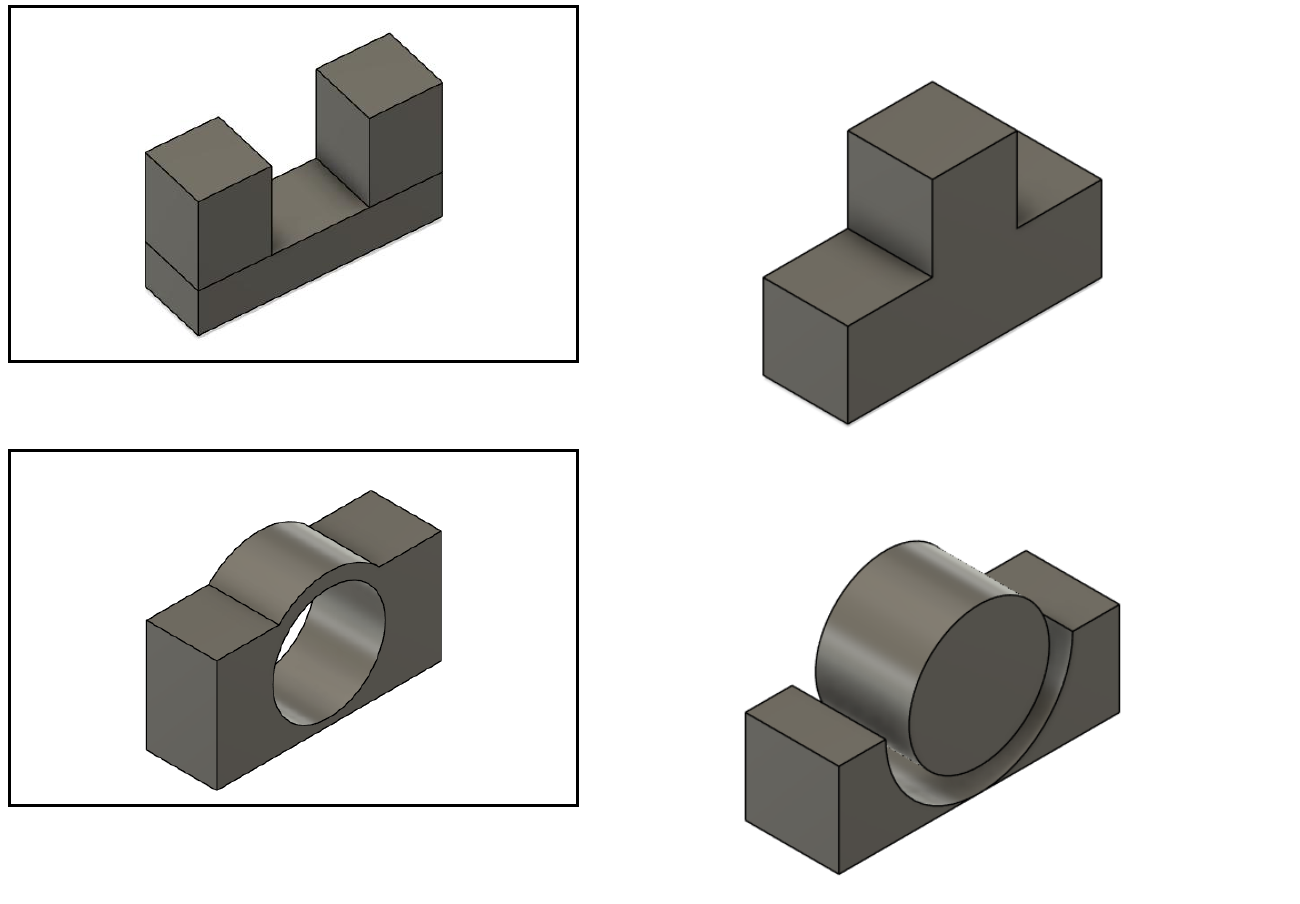}} & The difficulty in distinguishing concave and convex shapes results in erroneous models. This issue may arise because the vision tower can detect these shapes but cannot differentiate between them, or because the language model fails to generate the correct code, leading to sketch errors. \\ \hline
\end{tabular}
\caption{Error Cases and Related Analysis}
\label{tab:error_analysis}
\end{table*}

Based on the results and typical errors mentioned in the Table~\ref{tab:error_analysis}, we can infer the current shortcomings of the OpenECAD model:
\begin{itemize}
\item The model has weak handling capabilities for complex sketches. This includes issues such as the vision tower's low resolution, which leads to an inability to recognize shapes, and the language model's failure to correctly process complex shapes, resulting in erroneous sketch generation.
\item The generated models have poor size accuracy, particularly in handling the proportions between sketch dimensions and extrusion sizes. This results in mismatched parts, as the model fails to ensure consistent sizes between the sketches and extrusions of different components.
\item For complex shapes involving multiple ``Sketch-Extrusion'' steps, the model struggles to generate accurate results. It has difficulty remembering the previously drawn parts through the context, leading to errors.
\item Due to the current work focusing only on basic instructions without a combined instruction set like drawing rectangles in one command, the CAD operation codes for 3D shapes tend to be lengthy. Moreover, due to contextual limitations of the model, especially with OpenECAD 0.55B and 0.89B using OpenELM which supports only 2048 tokens, and considering that image inputs also consume tokens, there are situations where the code cannot be completed.
\end{itemize}

\subsection{An Example of Using OpenECAD models\label{section:exampleofmodel}}

\begin{figure*}[t]
    \centering
    \includegraphics[width=0.98\textwidth]{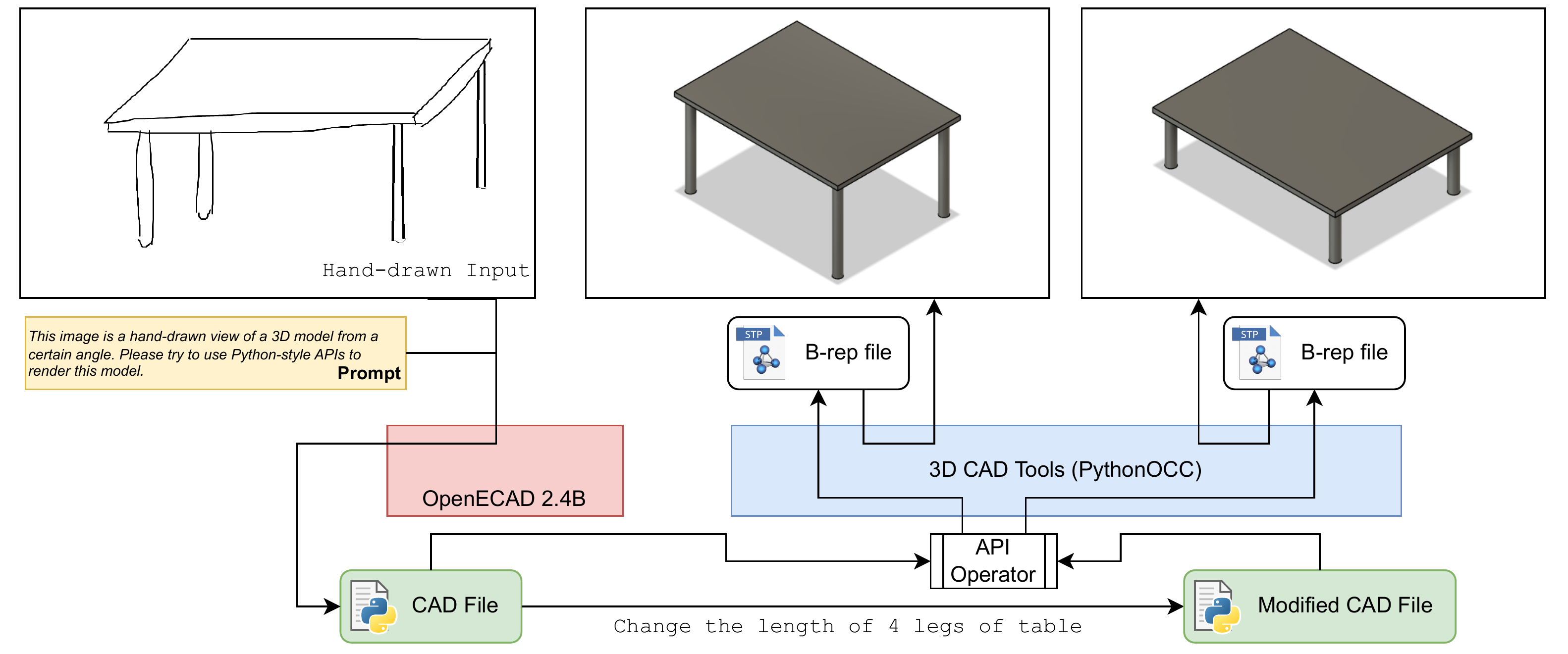}
    \caption{An example of using OpenECAD 2.4B model to draw a table.} %
    \label{fig:exampleofuse}
\end{figure*}

In this section, we will attempt to use the OpenECAD model by inputting a hand-drawn simple table (see Figure~\ref{fig:exampleofuse}), generating the corresponding OpenECAD code, changing the length of 4 legs of the table and then converting it into a step file using the PythonOCC API.

We input a hand-drawn table's photo to the OpenECAD 2.4B model, with the prompt:

\textit{``This image is a hand-drawn view of a 3D model from a certain angle. Please try to use Python-style APIs to render this model.''}

The model should respons:

\textit{``Of course, here are the codes:$<$OpenECAD code here$>$}

\begin{algorithm}[t]
    \caption{Some OpenECAD Code of the Table}	%
    \label{alg:example}
    \begin{algorithmic}[1]
        \STATE SketchPlane0 = add\_sketchplane(origin= [0., 0., 0.], normal= [0., 0., 1.], x\_axis= [1., 0., 0.])) 
        \STATE Loops0, Curves0\_0 = [], [] 
        \STATE Line0\_0\_0 = add\_line(start= [-1000.,  -750.], end= [1000., -750.])
        \STATE...
        \STATE SketchPlane1 = add\_sketchplane\_ref(
        Extrude0, origin = [0., 0.], type =  ``sameplan'', reverse = True)
        \STATE ...
        \STATE Extrude1 = add\_extrude(sketch= Sketch1,
        operation= 1, type= 0, extent\_one= 1000., extent\_two= 0.)
        \STATE \COMMENT{extent\_one is the length of one of 4 legs of the table.}
    \end{algorithmic}
\end{algorithm}

Some key code snippets are shown in the Algorithm~\ref{alg:example}. We changed the extent\_one of the table legs extrusion from 1000 to 500. We converted the code before and after the change into STEP files using a conversion tool and opened them with a CAD tool. As shown in the Figure~\ref{fig:exampleofuse}, the table legs are indeed shortened by half.

Additionally, besides directly modifying the OpenECAD code, it is also possible to convert the OpenECAD code into project files for other 3D CAD tools. For instance, AutoDesk Fusion supports all operations in OpenECAD. We can use tools and the AutoDesk Fusion API to convert OpenECAD code into AutoDesk Fusion project files. This way, we can fully utilize all the features of AutoDesk Fusion to continue modifying the design. Due to the limitations of the model's context length, it is currently challenging for the model to directly modify the generated OpenECAD code based on requirements through dialogue, as the context length does not support accommodating two complete OpenECAD code segments.

\subsection{Future Work\label{section:future_work}}
\subsubsection{Refinement}

Based on the above analysis, our future work will focus on the following areas:

\begin{itemize}
\item \textbf{Enhancing the dataset} involves several key strategies: expanding support for additional CAD instructions, annotating critical shape dimensions in input images to aid accurate size-based shape drawing by the model, selecting optimal angles for input images to showcase all features comprehensively, and employing diverse rendering tools to enhance the model's ability to recognize shapes under varied lighting conditions and artistic styles.
\item \textbf{Improving the visual model} is crucial as the current resolution limits its ability to distinguish between certain shapes. Therefore, training a visual model capable of handling higher resolutions is necessary. Additionally, integrating an image segmentation module like FastSAM\cite{zhao2023fast} before the visual model can aid in identifying which parts compose the 3D shapes.
\item \textbf{Improving the language model}'s capabilities is essential, as the current model supports a limited context length and has weak code generation abilities, resulting in incomplete or erroneous code. We can address this by using larger language models that support longer contexts, such as Phi-3-medium or LLaMA-3. Additionally, implementing Retrieval-Augmented Generation~(RAG) \cite{gao2023retrieval} and prompting techniques can further enhance the model's contextual understanding and code generation capabilities.
\item \textbf{Enhancing feedback integration with CAD tools}. By adopting a conversational approach, partly generated CAD instructions can be promptly sent to the CAD tool for rendering. The rendering results can then be immediately returned to the model, helping it understand the current drawing status, identify potential errors, correct them, and continue generating code based on the target shape and the already rendered results.
\end{itemize}

\subsubsection{Applications}

Currently, OpenECAD primarily serves as a simple CAD operation generation model, focusing on generating models from images. However, by improving the dataset, we can train models using similar methods and apply them to downstream applications. These applications include, but are not limited to:
\begin{itemize}
\item Assisting users in operating CAD design tools by guiding them on how to proceed with CAD operations to build or modify models based on their requirements and existing CAD actions.
\item Enhancing the ability to handle user CAD needs, such as design reusability and designing connectors for existing parts.
\item Introducing CAE-related knowledge to the model to address some CAE issues, such as structural reinforcement, material reduction, and kinematic simulation, during the CAD phase.
\end{itemize}

\section{Discussion and Conclusion\label{section:conclusion}}

For CAD model generation, our approach has some limitations. Currently, we only consider three of the most widely used curve command types (lines, arcs, and circles), but other curve commands can be easily added. We also only consider using a single image as a reference, with the model attempting to generate the complete output result rather than step-by-step outputs. Not every CAD command sequence code currently generates a topologically valid shape. Our network cannot guarantee the correctness of its output. In practice, when the context length is sufficient, the number of execution failures in the generated CAD command sequence code is relatively low. However, when the context length is insufficient, it becomes challenging to generate the code completely.

To address these limitations, more work needs to be done. Enhancements to the dataset should include support for multi-view perspectives, complex curves, and textual requirements. The code generation process should be improved to better integrate with CAD tools, allowing the model to generate code incrementally, similar to how humans draw while referencing the design, rather than outputting the entire CAD code at once.

In summary, we have introduced OpenECAD, a visual language model and its accompanying dataset designed for generating CAD operation sequence codes. OpenECAD aims to address the challenge of CAD model generation using visual language models and has successfully generated some relatively simple 3D shapes. We offer 4 sizes of OpenECAD models, ranging from 0.55B to 3.1B parameters.

\section*{CRediT authorship contribution statement}
\textbf{Zhe Yuan:} Software, Writing – original draft, Writing – review \& editing, Project administation. \textbf{Jianqi Shi:} Writing – review \& editing, Supervision. \textbf{Yanhong Huang:} Supervision.

\section*{Declaration of competing interest}
The authors declare that they have no known competing financial interests or personal relationships that could have appeared to influence the work reported in this paper.

\section*{Data availability}

Data will be made available on request.

\section*{Declaration of generative AI and AI-assisted technologies in the writing process}

During the preparation of this work the author(s) used ChatGPT in order to correct the possible wrong expression in English. After using this tool/service, the author(s) reviewed and edited the content as needed and take(s) full responsibility for the content of the published article.

\section*{Acknowledgment}

This research did not receive any specific grant from funding agencies in the public, commercial, or not-for-profit sectors.

\bibliographystyle{elsarticle-num} 
\bibliography{refs}
\end{document}